\documentclass[runningheads]{llncs}

\usepackage{eccv}

\usepackage{eccvabbrv}

\usepackage{graphicx}
\usepackage{booktabs}
\usepackage{multirow, bm}
\usepackage[accsupp]{axessibility}  % Improves PDF readability for those with disabilities.

\usepackage[pagebackref,breaklinks,colorlinks,citecolor=eccvblue]{hyperref}
\usepackage{orcidlink}

\begin{document}

\title{The Lottery Ticket Hypothesis in Denoising:\\
Towards Semantic-Driven Initialization} 

\titlerunning{The Lottery Ticket Hypothesis in Denoising}

\author{Jiafeng Mao\inst{1, 2} \and
Xueting Wang\inst{2} \and
Kiyoharu Aizawa\inst{1}}

\authorrunning{J.~Mao et al.}

\institute{
The University of Tokyo, Tokyo, Japan \email \\
{\{mao, aizawa\}@hal.t.u-tokyo.ac.jp} 
\and
CyberAgent, Tokyo, Japan  \email \\
{wang\_xueting@cyberagent.co.jp}}

\maketitle

\begin{abstract}
Text-to-image diffusion models allow users control over the content of generated images. Still, text-to-image generation occasionally leads to generation failure requiring users to generate dozens of images under the same text prompt before they obtain a satisfying result. We formulate the lottery ticket hypothesis in denoising: randomly initialized Gaussian noise images contain special pixel blocks (winning tickets) that naturally tend to be denoised into specific content independently. The generation failure in standard text-to-image synthesis is caused by the gap between optimal and actual spatial distribution of winning tickets in initial noisy images. To this end, we implement semantic-driven initial image construction creating initial noise from known winning tickets for each concept mentioned in the prompt. We conduct a series of experiments that verify the properties of winning tickets and demonstrate their generalizability across images and prompts. Our results show that aggregating winning tickets into the initial noise image effectively induce the model to generate the specified object at the corresponding location. Project Page: \url{https://ut-mao.github.io/noise.github.io}
\end{abstract}

\section{Introduction}
\label{sec:intro}

\begin{figure*}[t]
  \centering
  \includegraphics[width=1\linewidth]{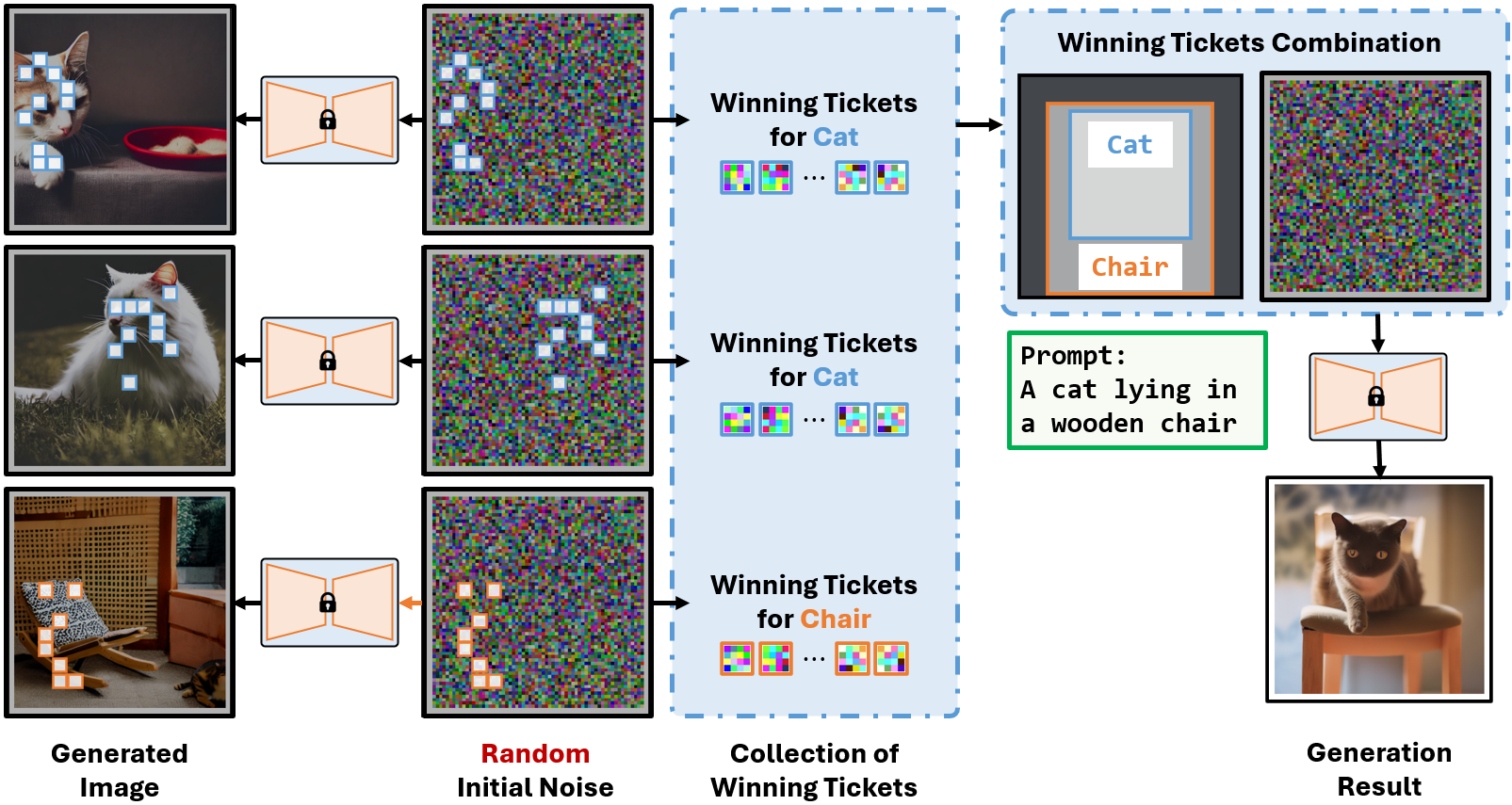}
  \caption{The concept of the winning ticket (special noisy pixel blocks) in denoising. These pixel blocks are naturally susceptible to being denoised to specific concepts. In this paper, we investigate and verify the property of winning tickets. We find that winning tickets collected from different images can be used to create new initial images and show significant advantages under different prompts.}
  \label{fig:1}
  \vspace{-10pt}
\end{figure*}

Diffusion models work by progressively denoising initial noise images drawn from a Gaussian distribution. Previous research in this area has primarily focused on the use of text prompts to guide image generation~\cite{rombach2022high, ramesh2022hierarchical, nichol2022glide, kim2022diffusionclip, ramesh2021zero, ding2021cogview}. While efforts have been made to enhance the denoising process or explore additional conditions for better control over the generation process~\cite{liu2022compositional, kawar2022imagic, feng2022training, hertz2022prompt, park2021benchmark, balaji2022ediffi}, generation failures can still occur. Especially prompts involving multiple objects or precise positional relationships often require users to generate numerous failed images from the same prompt before achieving a satisfactory result. 

In this paper, we unveil the existence of migratable pixel blocks in the initial random Gaussian noise that are independently associated with certain concepts. Intriguingly, our discovery mirrors the well-established Lottery Ticket Hypothesis in network pruning~\cite{frankle2018lottery}, suggesting the presence of a valuable subset within randomly initialized entities. Hence, we term our findings in this paper as the lottery ticket in denoising.

\vspace{5pt}
\noindent\textbf{The Lottery Ticket Hypothesis in Denoising.} \textit{Randomly-initialized noise images contain pixel blocks with initial values that allow them to be used to generate specific concepts.}
\vspace{5pt}

\noindent Specifically, during a standard generation process, when a user seeks to generate an image under a prompt $\mathcal{P}$ containing an object $c \in \mathcal{P}$, the randomly initialized image will posses special noisy pixel blocks, referred to as the "\textbf{winning tickets}", which can readily be denoised into $c$. Typically, these winning tickets are scattered randomly across the initial noise image at various locations. In the final generated image, $c$ will manifest at the location where its winning tickets are most concentrated. This hypothesis explains the reason for the occasional generation failure in standard text-to-image generation synthesis: When there is a significant gap between the spatial distribution of the winning tickets in the random initial image and the spatial distribution required for the content of the prompt description, the generation of the corresponding content fails. Our lottery ticket hypothesis predicts properties of the winning tickets as follows:
\begin{itemize}
    \item  Transplanting winning tickets from their original initial noise to a new noise image does not change the concept $c$ that these winning tickets can be easily denoised to.
    \item The winning tickets for concept $c$ still win under different prompts containing the same concept $c$.
\end{itemize}
In this study, we empirically confirm these properties and demonstrate that winning tickets can be extracted from the initial noise and used independently. The establishment of the first property demonstrates that the advantage of these winning tickets comes from their pixel values rather than from any external condition such as spatial location. The validity of the second property makes it possible to collect winning tickets without knowing the specific generation-time prompt. Combining both properties, we can first collect winning tickets from any prompt including $c$ and afterwards directly transplant them in noise images to generate new prompts containing $c$.

%that winning tickets can be extracted from the initial noise and used independently. In other words, winning tickets from different images can be combined to create new initial noise. In this study, we empirically confirm  that even transplanting these pixel blocks (winning tickets) from their original initial noise to a new noise image does not change the property that these winning tickets can be easily denoised to the corresponding concepts, offering evidence that  We also empirically confirm that 
%This property make it possible to collect winning tickets before knowing the specific generation-time prompt.

Diffusion models generate realistic images by iteratively refining random Gaussian noise, with current sampling methods typically requiring  dozens of denoising steps. In this study, we use a pre-trained diffusion model and identify winning tickets in the random initial noise as early as the first denoising step. Leveraging these identified winning tickets, we pioneer a semantics-driven approach to constructing initial noise, as shown in Fig.~\ref{fig:1}. Remarkably, \textbf{by merely substituting the initial noise image}, without any fine-tuning or disruption to the generation process, we achieve control over the placement of each concept. We also conduct experiments to demonstrate that methods guiding the denoising process yield significantly better results when applied to initial images composed of winning tickets based on our hypothesis.

In our experiments, we create initial noise by combining winning tickets, which does not guarantee that the initial image adheres to the Gaussian distribution. Surprisingly, directly employing pre-trained models on non-Gaussian initial noise still achieves high-quality images and the generated content follows the winning ticket location guidance. This result demonstrates that the diffusion model can tolerate non-Gaussian initial noise, eliminating the requirement for
exact adherence to the Gaussian distribution, thereby expanding the operational space for initial image manipulation.

The contributions of this paper can be summarized as follows:
\begin{itemize}
    \item We introduce the lottery ticket hypothesis in denoising as a novel finding and confirm the inter-image and inter-prompt invariance of winning tickets.
    \item We verify that the diffusion model can tolerate non-Gaussian initial images and highlight the importance of initialization, providing a new research direction for the interpretation and control of image generation.
    \item We present a straightforward yet powerful method for identifying and applying winning tickets in denoising, referred to as semantic-driven initial image construction.
    \item We demonstrate through experiments that initial images constructed by winning tickets consistently exhibit significant advantages, both independently and when combined with methods introducing denoising guidance.
\end{itemize}

\noindent\textbf{Implications.} Empirical investigation into the lottery ticket hypothesis in denoising provides valuable insights. By establishing the existence and properties of winning tickets, we hope to leverage this knowledge to:

\noindent\textbf{\textit{Improve fine-grained control.}} Given the significant impact of the spatial distribution of winning tickets in the initial image on the generation process, our findings emphasize the importance of considering the initial image when designing methods for achieving fine-grained control.

\noindent\textbf{\textit{For better theoretical understanding of diffusion models.}} The study of winning tickets offers a pathway to deepen our understanding of the denoising process, potentially benefiting research on diffusion model training and sampling acceleration.

\vspace{-10pt}
\section{Related Work}

Diffusion models have emerged as powerful tools for generating realistic images~\cite{dhariwal2021diffusion, ho2020denoising, ho2021classifier, nichol2021improved, song2020denoising, liu2022pseudo, li2023gligen}. Originating from pure Gaussian noise, these models iteratively predict the distribution of subsequent images, gradually refining noise samples until high-fidelity images are obtained. Many contemporary methods~\cite{kim2022diffusionclip, ramesh2021zero, ding2021cogview, gafni2022make} leverage text prompts projected by CLIP~\cite{radford2021learning} to guide the denoising process. Notably, stable diffusion~\cite{rombach2022high} conducts denoising in a latent space before decoding the latent representation into an image. Moreover, diverse approaches have been explored to introduce conditions enhancing the control over the generation process~\cite{liu2022compositional, mou2023t2i, kawar2022imagic, park2021benchmark, zhang2023adding, huang2023composer, wang2022pretraining, voynov2022sketch, avrahami2022blended, ruiz2023dreambooth, brooks2023instructpix2pix}, thereby broadening the utility of diffusion models. These endeavors predominantly focus on augmenting the controllability of diffusion models by refining the generation process.

In the realm of layout-to-image synthesis, generative models are tasked with placing specified objects in user-defined locations within prompts. Certain methodologies require training on annotated images to achieve this objective~\cite{cheng2023layoutdiffuse, zheng2023layoutdiffusion, jia2023ssmg, yang2023reco, xue2023freestyle, chen2023integrating, avrahami2023spatext}. Studies have noted that cross-attention maps closely align with the layout and structure of generated images~\cite{hertz2022prompt}, prompting exploration of control strategies based on manipulating these attention maps~\cite{feng2022training, balaji2022ediffi, mao2023trainingfree}. BoxDiff~\cite{xie2023boxdiff} propose to update the noise latent during the denoising under the layout guidance. But the initialization of noise is a step that has long been neglected. A recent work~\cite{mao2023guided} explores the generation tendency of initial images, but their discussion is limited to single images and only applicable when the prompt is known.

\section{Winning Tickets in Denoising}
\label{sec:proposal}

In this section, we delve into the cross-attention mechanism, which facilitates the identification of winning tickets in denoising (Sec~\ref{sec:attn}). Building upon our analysis, we present how we discover winning tickets and verify their properties. We construct a collection of lottery tickets, as illustrated in Sec.~\ref{sec:collect}, which facilitates winning ticket selection, as detailed in Sec~\ref{sec:select}. Our experiments are conducted using stable diffusion~\cite{rombach2022high}.

\subsection{Cross-Attention Layer and Winning Tickets}
\label{sec:attn}
\noindent\textbf{Cross-Attention Layer} The cross-attention maps can serve as a monitor displaying the similarity between each pixel block and each concept mentioned in the text prompt~\cite{mao2023guided}. Stable diffusion augments the U-Net~\cite{ronneberger2015u} with the cross-attention mechanism~\cite{vaswani2017attention} to allow for flexible condition. For the text-to-image generation task, a CLIP is employed to project text $p$ to a representation $\tau_\theta(p)$, which is subsequently mapped to the intermediate layers of the U-Net through a cross-attention layer as follows, 

\begin{equation}
    \text{Attention}(Q,K,V) = \text{softmax}(\frac{QK^T}{\sqrt{d}})\cdot{V}, 
    \label{eq:attention}
\end{equation}

\noindent where ${Q=W^{(i)}_Q\cdot{z_t}}, ~K=W^{(i)}_K\cdot\tau_\theta(p),~V=W^{(i)}_V\cdot\tau_\theta(p)$. The cross-attention layer computes the similarity between the query $Q$ (i.e., the embedding of the flattened input image $z_t$) and the key $K$ (i.e., the embedding of each word $p^{(i)}$ in the prompt $p$). Hence, the values in the attention map $M^{(i)}$ for a particular word $p^{(i)}$ indicate the corresponding pixel block's similarity with the word $p^{(i)}$. $V$ contains rich semantic features for each word $p^{(i)}$. 

\noindent\textbf{Winning Ticket}
In this study, we leverage attention maps from the most down-sampled cross-attention layer, which effectively preserves pixel block information~\cite{mao2023guided}. We partition each random noise image, sized $64 \times 64$, into $4 \times 4$ pixel blocks, resulting in a total of $256$ blocks per image. Each attention value on the map $M^{(i)}$ is determined by a $4 \times 4$ pixel block on the noisy image $z_t$ and a word embedding $p^{(i)}$, signifying the resemblance between this block and the concept $p^{(i)}$. Utilizing this attention map, the cross-attention layer reinforces features of $p^{(i)}$ into areas already resembling it, facilitating the denoising of pixel blocks initially exhibiting high cross-attention values for concept $p^{(i)}$. Pixel blocks demonstrating high cross-attention values on the map corresponding to word $p^{(i)}$ are identified as the winning tickets for concept $p^{(i)}$.

Overall, whether a pixel block is considered a winning ticket is determined by its relevance to the concepts. In other words, a pixel block may be considered a winning ticket for concept $c_1$ but not for concept $c_2$. The same pixel block may also become a winning ticket for multiple concepts, as long as it receives high cross-attention values for those concepts. In practical scenarios, when users intend to generate multiple objects, it's imperative to assign distinct winning tickets for each concept mentioned in the text prompt. This means that if a pixel block is identified as a winning ticket for both categories $c_1$ and $c_2$, and the user's desired generation scenario involves both categories, it's crucial to avoid using such ambiguous winning tickets. The reason behind this precaution is to prevent confusion in the denoising process, as utilizing such ambiguous winning tickets may hinder the model's ability to accurately determine whether to denoise the pixel block as category $c_1$ or $c_2$. To find winning tickets with significant tendency, we first collect a large number of noisy pixel blocks and design a selecting method for winning tickets.

\begin{figure*}[t]
  \centering
  \includegraphics[width=1\linewidth]{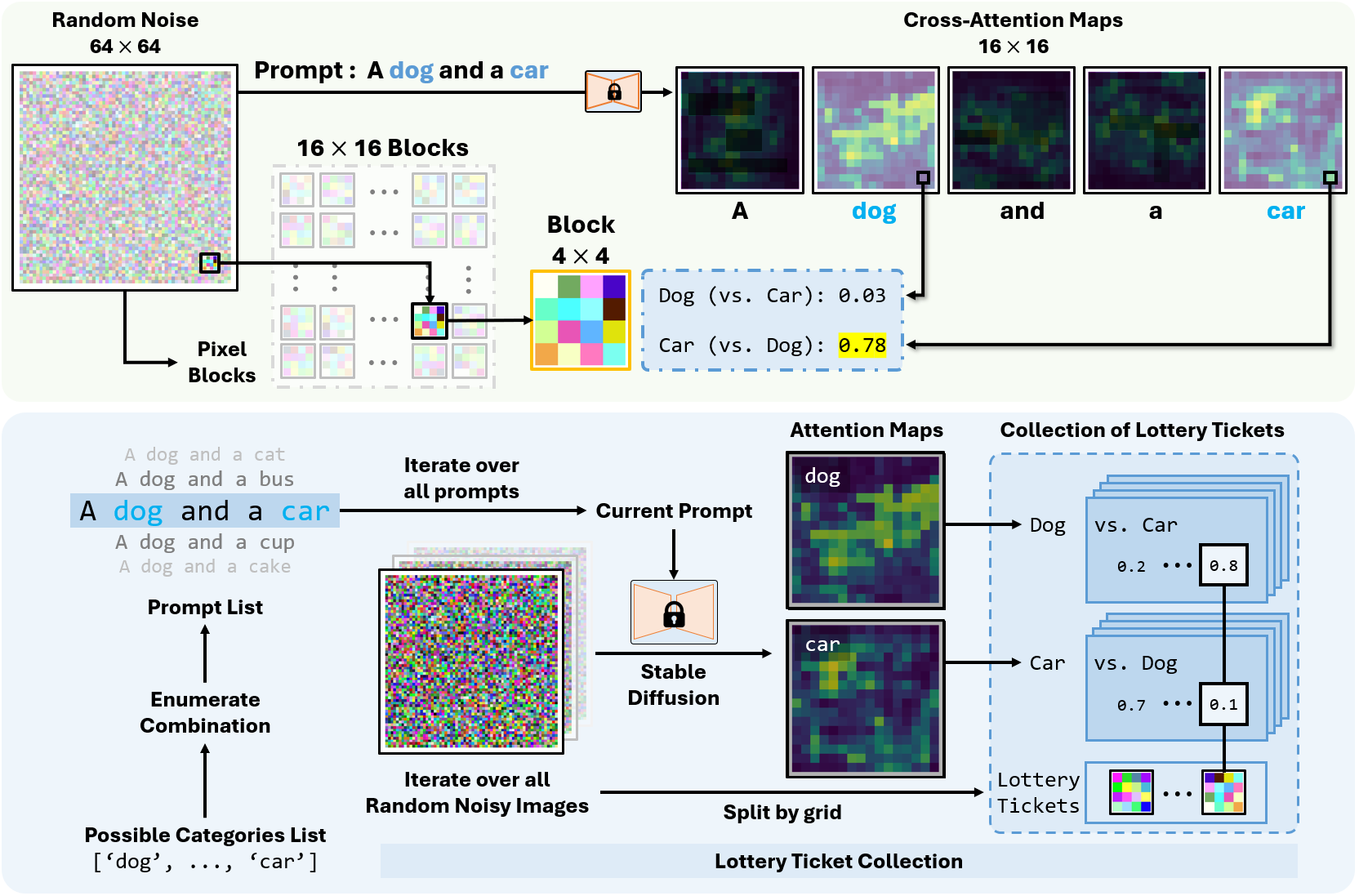}
  \caption{We create a collection containing a large number of pixel blocks and their scores on each possible category. The scores are extracted from the cross-attention maps calculated by the pre-trained stable diffusion. We will select winning tickets from this collection in the next phase.}
  \label{fig:collect}
  \vspace{-10pt}
\end{figure*}

\subsection{Lottery Tickets Collection}
\label{sec:collect}
In this section, we present a method to collect all pixel blocks (lottery tickets without knowing if they won) along with their category-wise similarity. 
To construct this database, a list of possible generation categories $\mathcal{L}$ needs to be specified in advance. Specifically, for the given list of possible categories $\mathcal{L}$, we enumerate all possible combinations of two different objects, $\{c_1, c_2\}$, to construct a list of prompts $\mathcal{L}_p$. The prompts take the form of \verb|[a |$c_1$ \verb|and a| $c_2$\verb|]|~(e.g. \verb|[a dog and a cat]|).

In our practical experience, we find that when the prompt mentions only one object (e.g., a photograph of a $c$), the pixel blocks with high attention values for the word $c$ are winning tickets for "foreground" rather than winning tickets for the object $c$, due to the absence of competition between different concepts. Therefore, in our experiments, we utilize contrastive prompts, i.e., prompts composed of a pair of objects (e.g., \verb|[a |$c_1$ \verb|and a| $c_2$\verb|]|), and perform a single step of denoising through the U-Net in the model. We collect the similarity information between tickets and these two concepts according to the cross-attention maps, as shown in the upper part of Fig.~\ref{fig:collect}. Since the attention values of pixel blocks are obtained through soft-max over words, obtaining a high score for word $c_1$ necessarily implies that this pixel block is a winning ticket for word $c_1$ over word $c_2$. We believe that such a design enables the screening of winning tickets with significant semantic information.

As we illustrated in Sec.~\ref{sec:attn}, whether a pixel block is considered a winning ticket is determined by its relevance to the corresponding concept. In each generation under the prompt containing different categories, different pixel blocks become winning tickets. For simplicity, we collect all lottery tickets from $\mathcal{N}$ noisy images $\mathcal{Z} = \{z_1, z_2, ..., z _\mathcal{N}\}$ sampled from the Gaussian distribution into a collection of lottery tickets. As shown in Fig.~\ref{fig:collect}, for each noisy image $z_i \in \mathcal{Z}$, we iterate through all prompts $p_j \in \mathcal{L}_p$. Utilizing the combination $\{z_i, p_j\}$, we only compute the cross-attention maps $\{\mathcal{M}_{c_1}, \mathcal{M}_{c_2}\}$ for the first step of denoising, where $\{c_1, c_1\} \subset p_j$. The final cross-attention values are calculated via the softmax function across all words within the prompt~\cite{vaswani2017attention}. Each pixel block $b^{k}_i \in z_i$ is logged into the collection, along with its score on $\{\mathcal{M}_{c_1}, \mathcal{M}_{c_2}\}$ in the entry of $c_1$ (\verb|vs.| $c_2$) and $c_2$ (\verb|vs.| $c_1$), respectively. Such an operation is performed on all combinations $\{z, p~|~z \in \mathcal{Z}, p \in \mathcal{L}_p \}$. For simplicity, we use $\mathcal{S}_{1,(2)}$ to denote the \textbf{$\mathcal{S}$}core entry of $c_1$ (\verb|vs.| $c_2$) in the collection. A high score in $\mathcal{S}_{1,(2)}$ implies that the corresponding lottery ticket can be a winning ticket for $c_1$  when $c_1$ and $c_2$ both appear in the prompt. For each lottery ticket $b$ in the collection, we calculate an average score for each category $c_i$ by averaging across all combinations of $c_i$ (\verb|vs.| $c_j$), where $j \neq i$, as follows,
\begin{equation}
\vspace{-5pt}
    \text{Average Score}_{i} = \frac{1}{\mathcal{N}_c-1} \sum_{j \neq i} \mathcal{S}_{i,(j)}
    \label{eq:avg}
\end{equation}
where $\mathcal{N}_c$ indicates the number of possible categories.  

% Although we collect scores by category pairs, our method can handle \textbf{prompts containing more than two objects} during the generation. We will explain it in detail in Eq.~\ref{eq:filter}.

\subsection{Winning Tickets Selection}
\label{sec:select}

We select winning tickets based on our constructed collection of lottery tickets, which contains a diverse set of pixel blocks, each associated with scores for different category pairs. A high score in $c_1$ (\verb|vs.| $c_2$) indicates that the corresponding pixel block is well-suited for generating $c_1$ when $c_2$ is also present in the prompt. we can select winning tickets for any list of categories $\mathcal{C}=\{c_i, c_2, ..., c_n\} \subseteq \mathcal{L}$. Specifically, we select winning tickets $\mathcal{B}_i$ for $c_i$ using the following criterion:
\begin{equation}
    \mathcal{B}_i = \{ b~|~\forall{c_j \in \mathcal{C}, j\neq i}, \mathcal{S}_{i,(j)}(b) > \mathcal{T}_{obj}\},
    \label{eq:filter}
\end{equation}
where $\mathcal{S}_{i,(j)}(b)$ denotes the score in the entry of $c_i$ (\verb|vs.|$c_j$) corresponding to the lottery ticket $b$, and $\mathcal{T}_{obj}$ denotes a predefined threshold. 

Although we collect scores by category pairs, we experimentally demonstrate that using \textbf{prompts containing more than two objects} during the generation does not affect the property of the winning tickets we collect, as shown in Sec.~\ref{sec:otf}. In the case where the given list of categories $\mathcal{C}$ only contains a single category $c_i$, we extend the utilization of the average score to select winning tickets specifically tailored for that singular object $c_i$, as follows,

\vspace{-5pt}
\begin{equation}
\vspace{-5pt}
    \mathcal{B}_i = \{ t~|~\text{Average Score}_{i}(b) > \mathcal{T}_{obj}\},
    \label{eq:1obj}
\end{equation}
where $\text{Average Score}_{i}(b)$ denotes the average score of the pixel block $b$ on the category $c_i$.

We select pixel blocks with low average scores for all categories $c_i \in \mathcal{C}$ to obtain winning tickets for the background. 
\begin{equation}
    \mathcal{B}_{bg} = \{ b~|~\forall{c_i \in \mathcal{C},} \text{Average Score}_{i}(b) < \mathcal{T}_{bg}\},
    \label{eq:bg}
\end{equation}
where $\mathcal{T}_{bg}$ is a predefined threshold. 

\section{Property Verification of Winning Tickets}
\label{sec:exp}
We present the most straightforward application of the collected winning tickets in Sec.~\ref{sec:apply}, i.e., constructing regions in the initial noise by the winning tickets to induce the model to generate the corresponding concepts in these regions. We present two distinct usages and their results in Sec.~\ref{sec:otf} and Sec.~\ref{sec:adv}, which lead to interesting insights and analysis, as detailed in Sec.~\ref{sec:discussion}.

\subsection{Semantic-Driven Initial Image Construction}
\label{sec:apply}
Following the process we introduced in Sec.~\ref{sec:proposal}, given a category list $\mathcal{L}$, we can obtain winning tickets for each concept $c \in \mathcal{L}$. This process has two distinct usage scenarios as follows:
\begin{itemize}
    \item On the fly: After the user has given the target generation prompt, our method can immediately create a collection based on the list $\mathcal{L}$ of categories extracted from the prompt and select the winning tickets from the collection before generating the image (Sec.~\ref{sec:otf}). 
    \item In advance: Our method can create a collection based on a list $\mathcal{L}$ containing a large number of possible categories in advance, so that winning tickets can be selected directly from this collection when generating images.(Sec.~\ref{sec:adv}).
\end{itemize}

\begin{figure*}[t]
  \centering
  \includegraphics[width=1\linewidth]{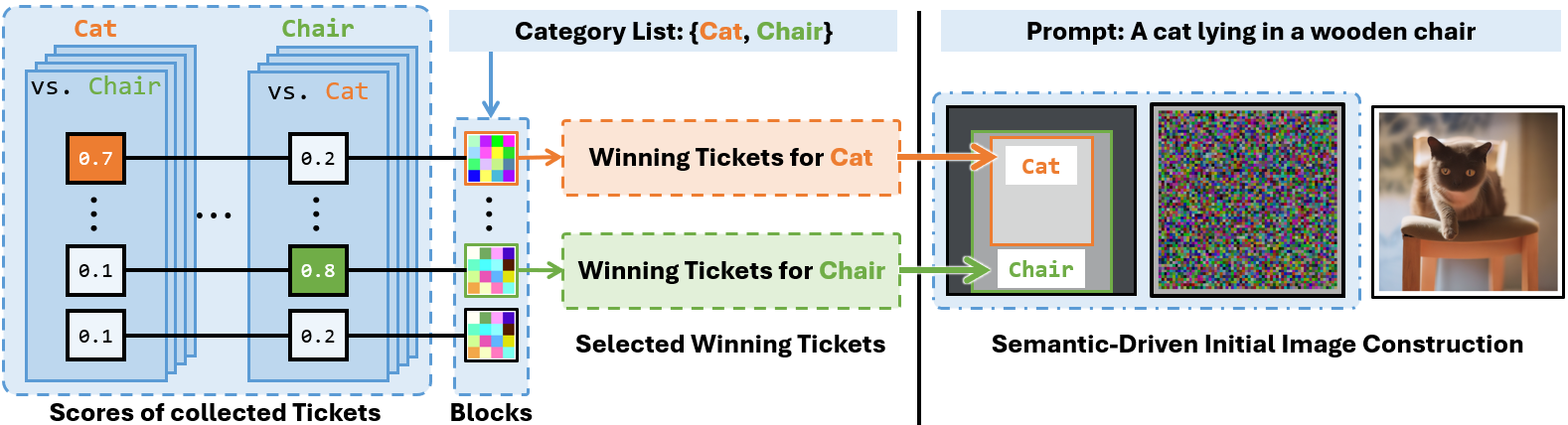}
  \vspace{-10pt}
  \caption{The collection contains scores for all lottery tickets, contrasting each category against others ($c_i$ (vs. $c_j$)). These scores serve as critical metrics used in the selection of the winning ticket.}
  \vspace{-10pt}
  \label{fig:select}
\end{figure*}

\begin{figure*}[t]
  \centering
  \includegraphics[width=1\linewidth]{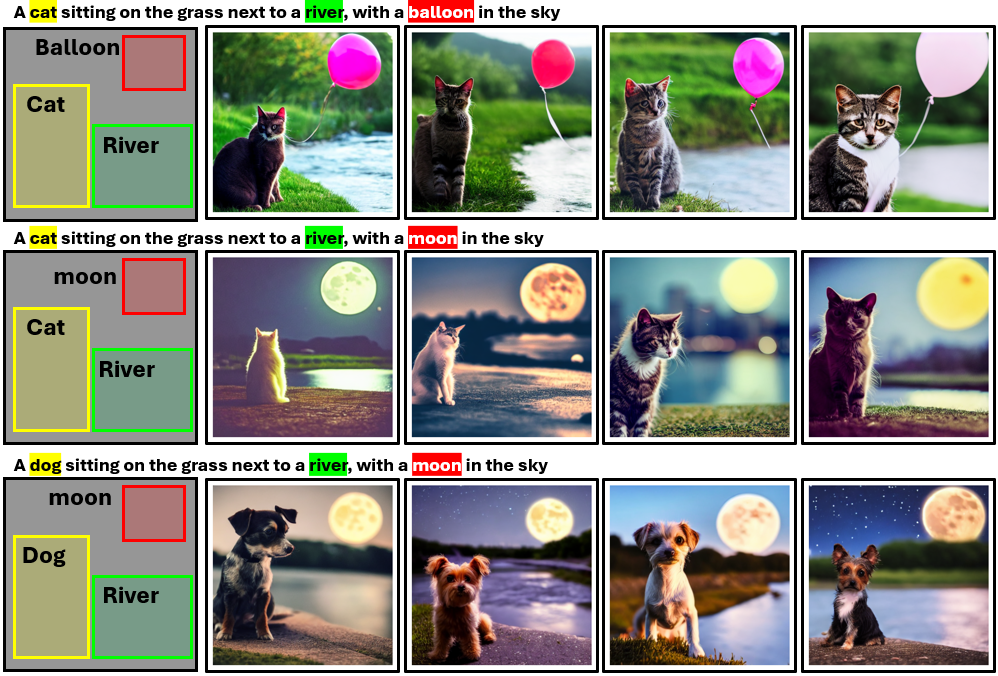}
  \caption{Quality samples in our on-the-fly experiments, under the same specified regions and different prompts. By merely replacing the constructed initial noise image, without interfering with the generation process, the model spontaneously generates the specified concepts in regions consisting of the winning tickets corresponding to these concepts after dozens of denoising steps. }
  \label{fig:3_obj}
  \vspace{-10pt}
\end{figure*}

\begin{figure*}[t]
  \centering
  \includegraphics[width=1\linewidth]{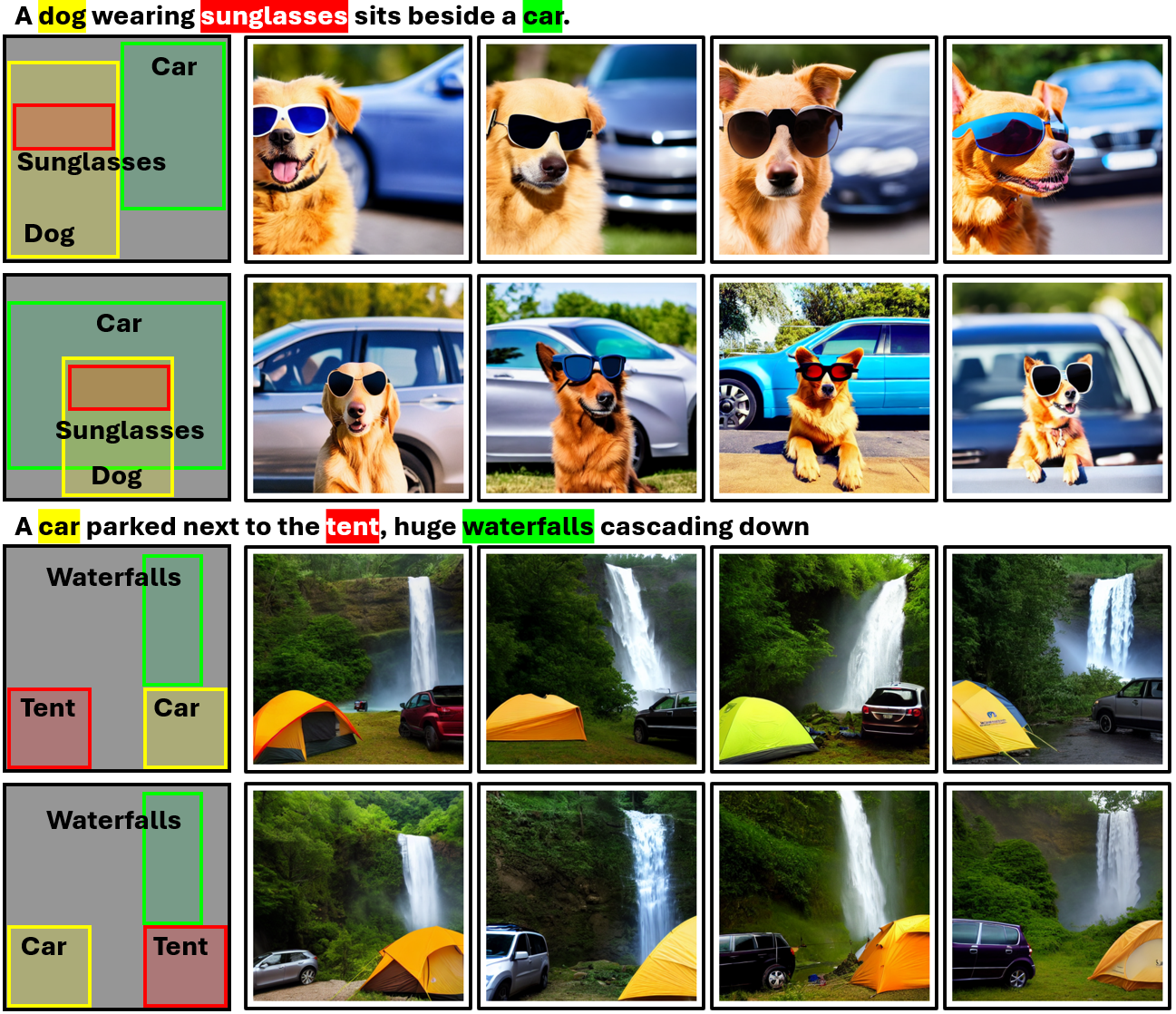}
  \caption{Results using the same prompt and different specified regions. The region composed of the winning tickets effectively induces the model to generate objects in it.}
  \vspace{-10pt}
  \label{fig:3_sunglasses}
\end{figure*}

\begin{figure*}[t]
  \centering
  \includegraphics[width=1\linewidth]{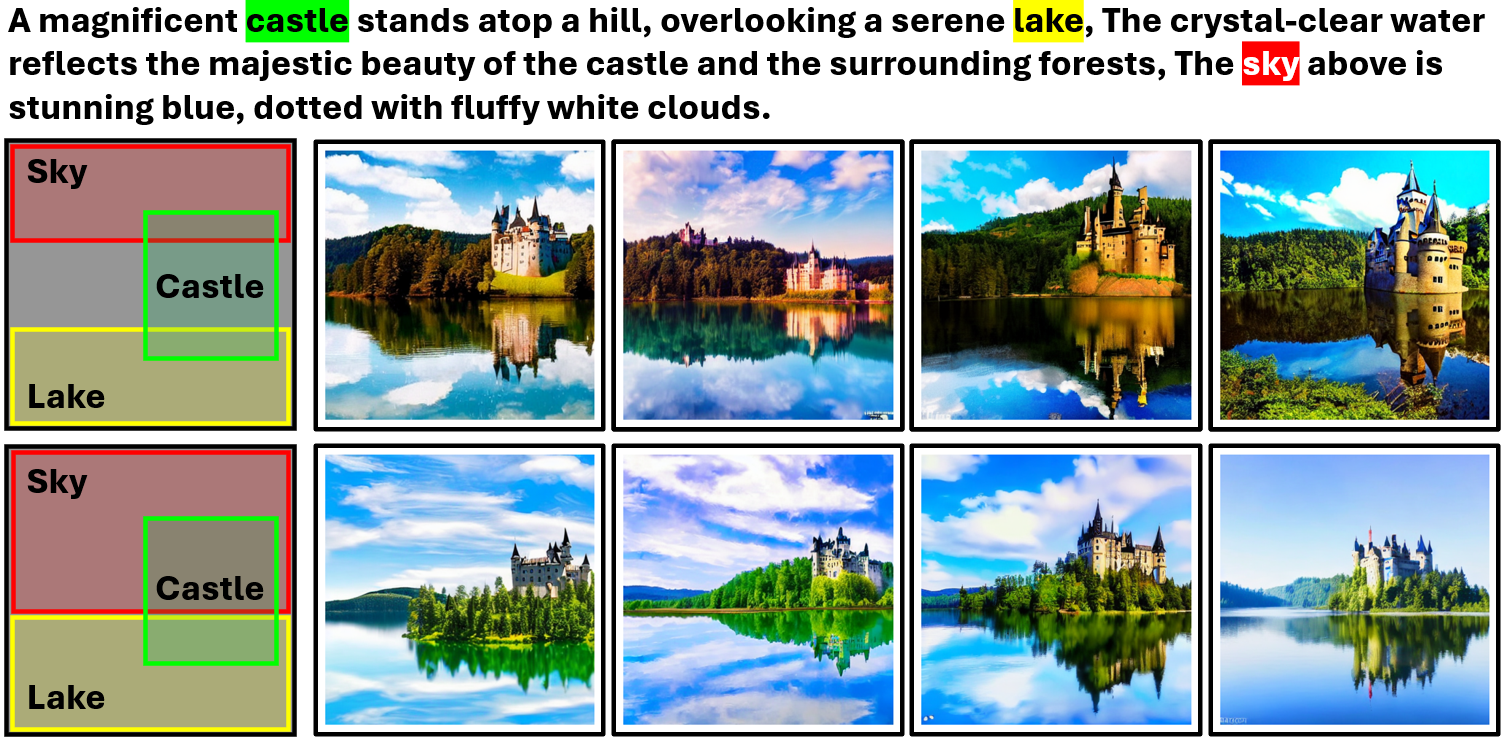}
  \caption{Results using the same prompt and different specified background regions.}
  \label{fig:3_castle}
  \vspace{-10pt}
\end{figure*}

We present a straightforward strategy of ultilizing these selected winning tickets. Specifically, for a prompt and specified locations of each object given by the user, we construct the corresponding area in the initial image by the winning tickets for the corresponding concept, as illustrated in Fig.~\ref{fig:select}. We randomly select the number of pixel blocks needed to construct the specified region from the filtered winning tickets and randomly fill the corresponding region with these selected winning tickets. We refer to this process as \textbf{Sem}antic-driven \textbf{I}nitialization (SemI) as we construct the initial noise using winning tickets with recognized semantic information. After that, we use the pre-trained diffusion model to denoise our constructed initial noise normally. When the number of winning tickets available is fewer than the number needed, we randomly reuse the winning tickets. We utilize the default configuration of SD~\cite{rombach2022high} v1.4 for the generation. In this setting, the model undergoes 50 denoising sampling steps using DDIM~\cite{song2020denoising}.

\vspace{-10pt}
\subsection{Collection On the Fly}
\label{sec:otf}
In this section, we manually create prompts containing multiple objects and manually specifying the locations of the individual objects to confirm the effectiveness of winning tickets in application. After we are given the prompt $p$, we use the set of objects contained in the prompt as the possible category list $\mathcal{L}$ and select the winning tickets from multiple random noise images. As shown in Fig.~\ref{fig:3_obj}, and \ref{fig:3_sunglasses}, selected winning tickets proficiently direct the generation of the three specified objects. Moreover, if descriptive words related to the background, such as \verb|sky|, \verb|lake|, etc., are included in the categories list during the collection creation, we can also find winning tickets for these background elements. In Fig.~\ref{fig:3_castle}, the top and bottom halves utilize identical prompts, differing only in the designated background regions. This difference is prominently reflected in the denoising results, demonstrating the power of winning tickets. More visual samples are provided in the supplementary.

The collection creation incurs a time cost. Let $\mathcal{N}_p$ denote the number of possible category pairs and the number of initial images is $\mathcal{N}$, the time cost of creating the collection is $\mathcal{N}_p \cdot \mathcal{N} \cdot t$, where $t$ represents the time required for a single denoising step. For stable diffusion v1.4 on an NVIDIA V100, $t$ is approximately $0.15$ seconds. For instance, when the given prompt includes 3 objects, creating a collection with $\mathcal{N}=10$ takes roughly 5 seconds, and $\mathcal{N}=100$ requires approximately 1 minute. The time cost of the winning ticket selection and the construction of the initial image are negligible.

\vspace{-10pt}
\subsection{Collection in Advance}
\label{sec:adv}
To verify the behavior of winning tickets in the presence of a large number of object combinations, we conducted an experiment where winning tickets were collected in advance against a large number of object pairs. We conduct this experiment on the MS COCO 2017 dataset~\cite{lin2014microsoft}, which contains both image captions and object annotations. The captions of images can be used as prompts, and the object annotations can be used as specified regions of objects. We select 6000 samples whose bounding box categories are all mentioned in their image captions, where 3000 contain one object, and the other 3000 contain two. There are a total of about 700 possible category combinations in the samples.

\vspace{-10pt}
\subsubsection{Control Strength Evaluation}
To quantitatively evaluate the effectiveness of winning tickets, we use an object detector~\cite{wang2023you} to perform object detection on generated images and obtain the categories and detected bounding boxes of generated objects~\cite{mao2023trainingfree}. Then the consistency of the detection results with the layout guidance can represent the effectiveness of our controlling. For each object, the Intersection over Union (IoU) between its layout guidance bounding box (i.e., object annotation in the dataset) and detected bounding box is calculated. If the IoU is larger than $0.5$, it is regarded as a successful control, and the success rate is represented by $R_\text{suc}$. To evaluate the effectiveness of objects of different sizes, all objects are categorized into several subsets according to their size. The subset $s$ contains objects with an area less than $150^2$ pixels, the subset $l$ contains all objects with an area greater than $300^2$ pixels, while the other objects are channeled into subset $m$. 

\vspace{-20pt}
\subsubsection{Implementation}
Stable diffusion pre-trained for the text-to-image task does not receive any layout guidance, but may by chance generate the specified objects in the specified regions. For experiments on other methods, the model receives both prompts and layout guidance and produces corresponding images. Our method uses the same pre-trained model to create our collection. The results of our method reported in Tab.~\ref{tab:results} are obtained under the default hyper-parameters of $\{\mathcal{N}, \mathcal{T}_{obj}, \mathcal{T}_{bg}\} = \{100, 0.5, 0.1\}$. An analysis of hyper-parameters is provided in Sec.~\ref{sec:hyper}. In this experiment, we create a collection based on all possible pairs of categories ($\sim700$) in the dataset. This results in approximately $3$ hours to create the collection of lottery tickets.

\vspace{-10pt}
\subsubsection{Combination with related methods}
Some training-free mask-based methods~\cite{balaji2022ediffi, mao2023trainingfree} are proposed for the layout-to-image synthesis task. We compare our method to them, and also since these methods focus on providing additional guidance merely through adding masks on cross-attention maps during the denoising process, they are fully orthogonal to what is discussed in this paper. We, therefore, try to use our method in conjunction with these methods to confirm that initialization using winning tickets improves the performance of these related methods. A recent work~\cite{mao2023guided} proposes to manipulate initial noise inside only a single random noise image but suffers from the limited number of winning tickets. Therefore, we also try to improve its performance by combining them with ours.

\vspace{-10pt}
\subsubsection{Results}
As shown in Tab.~\ref{tab:results}, our method achieves better performance than the Paint~\cite{balaji2022ediffi} and Swap~\cite{mao2023guided}, and comparable performance with Soft~\cite{mao2023trainingfree}, in terms of control effectiveness, and combining with ours significantly boosts the performance of all other methods. This experimental result exemplifies the superiority of using winning tickets to construct the initial noise.

\vspace{-10pt}
\subsubsection{Hyper Parameters}
\label{sec:hyper}
We conducted a quantitative sensitivity analysis by varying parameters within the set $\{\mathcal{N}, \mathcal{T}_{obj}, \mathcal{T}_{bg}\}$ and observed the resulting changes, as depicted in Tab.~\ref{tab:hype}. Larger values of $\mathcal{T}_{obj}$ and smaller values of $\mathcal{T}{bg}$ improve the control over small and medium-sized objects. However, this heightened control leads the value distribution of winning tickets to severely deviate from Gaussian distribution, thereby leading to worse performance for control of large-size objects, where more winning tickets are utilized to construct the initial noise.

Our strategy of reusing winning tickets, as illustrated in Sec.~\ref{sec:apply}, is influenced by $\mathcal{N}$. A smaller $\mathcal{N}$ implies fewer pixel blocks in the collection, causing the constructed image to be more likely to contain duplicate pixel blocks, especially for control of large-size objects. Thus, a larger $\mathcal{N}$ yields better image quality.

\begin{table*}[t]
  \begin{center}
    \setlength{\tabcolsep}{1mm}{
  \begin{tabular}{l|cccc|cccc}
    \hline
     Methods & IoU $\uparrow$  &  $\text{IoU}_s\uparrow$  & $\text{IoU}_m\uparrow$ & $\text{IoU}_l\uparrow$ & $R_{\text{suc}}\%\uparrow$ & $R^s_{\text{suc}}\%\uparrow$ &$R^m_{\text{suc}}\%\uparrow$ &$R^l_{\text{suc}}\%\uparrow$    \\
    \hline
    \hline
    SD~\cite{rombach2022high} & $0.19$ & $0.05$ & $0.21$ & $0.38$ & $12.14$ &$ 0.50$ & $8.11$ & $39.41$ \\
    Paint~\cite{balaji2022ediffi}  &  $0.24$ & $0.08$ & $0.27$ & $0.46$ & $17.56$ & $1.16$ & $14.05$ & $52.34$ \\
    Soft~\cite{mao2023trainingfree} & $0.27$ &  $0.10$  & $0.31$  &  $0.48$ &   $22.31$ &  $2.95$  &  $22.38$ &  $56.24$ \\
    Swap~\cite{mao2023guided} & $0.26$ & $0.09$ &  $0.32$ & $0.46$ & $21.58$ & $2.26$ &  $22.75$ & $54.21$\\
    \hline
    \hline
    $\text{SemI}_\text{(ours)}$ & $0.26$ & $0.10$ & $0.32$ & $0.45$ & $21.94$ & $2.40$ & $24.68$ & $52.31$ \\
    SemI + \scalebox{0.7}{Soft~\cite{mao2023trainingfree}} & $0.31$ & $0.14$ & $0.38$ & $0.50$ & $28.73$ & $5.45$ & $33.49$ & $62.34$\\
    SemI + \scalebox{0.7}{Swap~\cite{mao2023guided}} & $0.28$ & $0.11$ &  $0.34$ & $0.46$ & $23.73$ & $3.38$ &  $27.25$ & $54.21$ \\
    SemI + \scalebox{0.7}{\cite{mao2023trainingfree} \& \cite{mao2023guided}} & \bm{$0.32$} & \bm{$0.15$} & \bm{$0.39$} & \bm{$0.52$} & \bm{$29.29$} & \bm{$5.73$} & \bm{$34.42$} & \bm{$62.74$}\\
    \hline
  \end{tabular}}
\end{center}
\caption{Constructing initial noise solely through the utilization of winning tickets achieves a control effectiveness comparable to existing layout-to-image methods, and incorporating winning tickets alongside them significantly enhances their performance further.}
\label{tab:results}
\vspace{-10pt}
\end{table*}

\begin{table*}[t]
  \begin{center}
    \setlength{\tabcolsep}{1.2mm}{
  \begin{tabular}{l|cccc|cccc}
    \hline
     Parameters & IoU $\uparrow$  &  $\text{IoU}_s\uparrow$  & $\text{IoU}_m\uparrow$ & $\text{IoU}_l\uparrow$ & $R_{\text{suc}}\%\uparrow$ & $R^s_{\text{suc}}\%\uparrow$ &$R^m_{\text{suc}}\%\uparrow$ &$R^l_{\text{suc}}\%\uparrow$   \\
    \hline
    Default & $0.26$ & $0.10$ & $0.32$ & $0.45$ & $21.94$ & $2.40$ & $24.68$ & $52.31$ \\
    \hline
    \hline
    $\mathcal{N} = 10$ & $0.25$ & $0.08$ & $0.31$ & $0.44$ & $21.76$ & $2.10$ & $25.00$ & $51.45$ \\
    $\mathcal{N} = 50$ & $0.26$ & $0.09$ & $0.32$ & $0.45$ & $22.98$ & $3.05$ & $26.31$ & $53.01$ \\
   
    \hline
     $\mathcal{T}_{obj} = 0.3$  & $0.25$ & $0.08$ & $0.31$ & $0.45$ & $20.32$ & $1.76$ & $21.91$ & $50.90$ \\
     $\mathcal{T}_{obj} = 0.7$ & $0.25$ & $0.10$ &  $0.32$ & $0.41$ & $21.32$ & $3.05$ &  $25.12$ & $47.59$ \\
    \hline
     $\mathcal{T}_{bg} = 0.05$  & $0.26$ & $0.10$ &  $0.32$ & $0.45$ & $22.36$ & $3.02$ &  $24.97$ & $52.56$ \\
     $\mathcal{T}_{bg} = 0.2$  & $0.25$ & $0.08$ &  $0.31$ & $0.45$ & $21.19$ & $2.10$ &  $22.55$ & $53.11$ \\
    \hline
  \end{tabular}}
\end{center}
\caption{Quantitative sensitive analysis over all hyper-parameters. The default setting is $\{\mathcal{N}, \mathcal{T}_{obj}, \mathcal{T}_{bg}\} = \{100, 0.5, 0.1\}$.}
\vspace{-10pt}
\label{tab:hype}
\end{table*}

\vspace{-10pt}
\subsection{Discussion}
\label{sec:discussion}
The successful generation of images and the ability to exert effective control through utilizing winning tickets are profoundly enlightening for \textbf{three significant reasons} as follows:

\noindent\textbf{Tolerance for non-Gaussian Initial Images} All of the initial images used for denoising in our methods are constructed and do not adhere to the Gaussian distribution. According to our results (Fig.~\ref{fig:3_obj}, ~\ref{fig:3_sunglasses} and Tab.~\ref{tab:results}), most of the images generated from non-normal initial images still exhibit high quality while being effectively guided. However, the diffusion model can tolerate non-Gaussian initial images to a certain extent, but the generation still fails when the initial image deviates extremely from the normal distribution.

\begin{figure*}[t]
  \centering
  \includegraphics[width=1\linewidth]{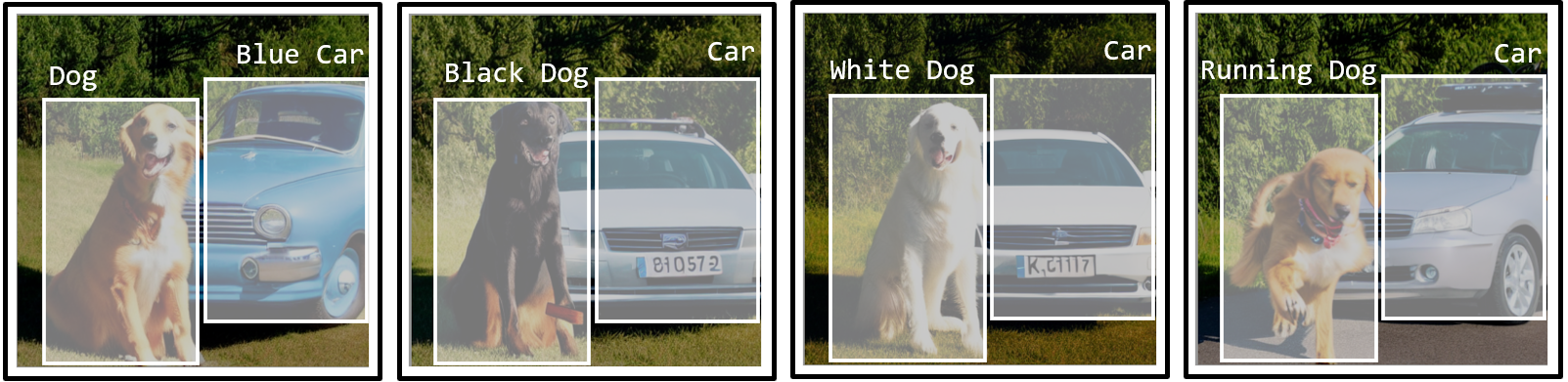}
  \caption{Versatility of winning tickets over different prompts containing the same objects. These four images are generated from the same constructed initial image and \textbf{different prompts}. The pixel blocks used for constructing each area are selected according to their scores on the simple prompt \textbf{a dog and a car}. The generation tendency of those pixel blocks remains unchanged even with different descriptions, such as \textbf{black dog} or \textbf{running dog}. }
  \label{fig:ver}
\end{figure*}

\noindent\textbf{Versatility of Winning Tickets over Prompt} In our experiment, we create a collection in advance by simply accepting a list of possible categories, without knowing the specific prompts that the generation uses. We adopt a simplistic prompt format~(e.g. \verb|[a dog and a cat]|) to select the winning tickets. According to our evaluation, winning tickets evaluated under such simple prompts can generate objects in various prompts efficiently. Our results reveal that a winning ticket has considerable versatility between different prompts containing the same objects. Winning tickets of \verb|dog| under our simple prompt \verb|[a dog and a car]| demonstrate applicability across various prompt descriptions, such as \verb|running dog| or \verb|black dog|, as shown in Fig.~\ref{fig:ver}.

\noindent\textbf{Versatility of Winning Tickets over Image} Our results confirm the versatility of winning tickets across different initial images.  Whether a pixel block can be a winning ticket is determined by its initial pixel values. winning tickets from one image will win again when transferred to another image.

\vspace{-10pt}
\subsection{Limitations and Future Works} 

\begin{figure*}[t]
  \centering
  \includegraphics[width=1\linewidth]{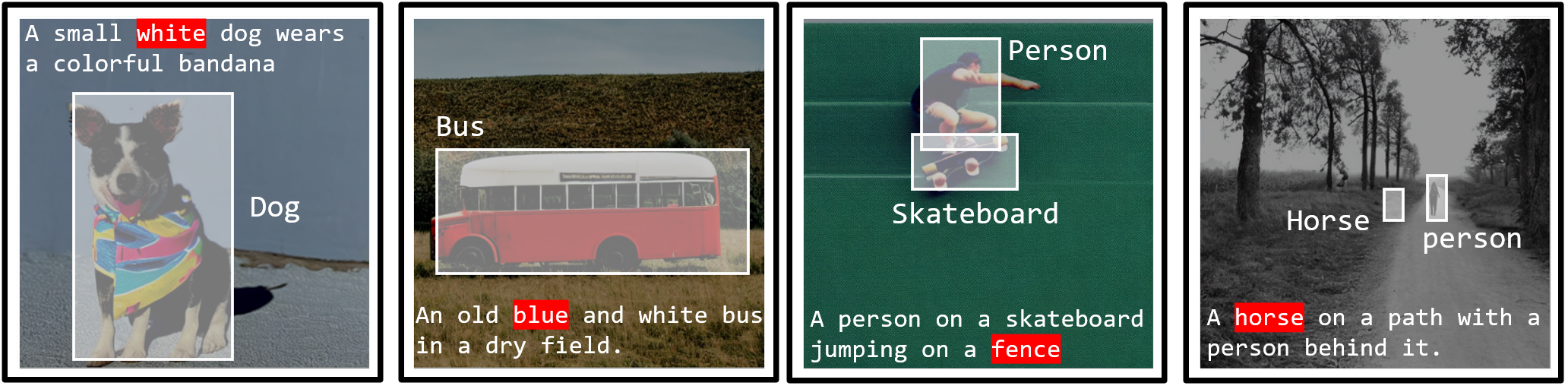}
  \caption{Typical failure cases in our experiment. Each image is generated from a constructed noise according to the layout guidance denoted by the white layout guidance.}
  \label{fig:fail}
  \vspace{-10pt}
\end{figure*}

For clarity of explanation, we use typical failure cases to illustrate the limitations of our approach of utilizing winning tickets, as shown in Fig.~\ref{fig:fail}. In the first two images, both the dog and the bus have the wrong colors. The winning ticket selection in our method is based solely on the names of the categories. Occasionally, these selected winning tickets may exhibit a tendency to generate certain random properties (e.g. \verb|red| bus), and such tendencies overcome the prompt during the generation. This can probably be addressed by a finely designed method of selecting winning tickets considering other words in the prompt. 

The position control of the third image is successful, but the quality of this image is compromised and the background generation fails. As a small number of the initial noise constructed by winning tickets deviates extremely from the normal distribution, images generated from them carry certain unnatural features. Recalibrating the constructed initial image to adhere to a normal distribution may alleviate this issue.  

The last image contains two very small objects, and our method only succeeds in generating the person. Although our method significantly outperforms other methods on control over small objects, when the guidance box is extremely small, the generation is still challenging.

\vspace{-10pt}
\section{Conclusion}

This work introduces and validates the Lottery Ticket Hypothesis in Denoising for text-to-image generative diffusion models, revealing that certain pixel blocks within initial noise images—termed \textit{winning tickets}—possess inherent predispositions towards generating specific content. Our findings illuminate a previously underexplored aspect of the denoising process and propose a novel, semantic-driven approach for initial image construction that significantly enhances control over the generated imagery. By empirically demonstrating the existence, properties, and applicability of winning tickets across various images and prompts, we have opened new pathways for improving the precision and flexibility of image generation. This research not only advances our understanding of diffusion models but also sets the stage for further explorations into optimizing generative processes for more deterministic and customizable outputs.

\clearpage  % TODO REVIEW/FINAL: This \clearpage needs to be removed from both review and camera-ready versions.

\bibliographystyle{splncs04}
\bibliography{main}
\end{document}